\documentclass[final]{l4dc2026}

\title[Nuclear Reactor Core Simulation through Surrogate Models]{Enhancing Nuclear Reactor Core Simulation through Data-Based Surrogate Models}
\usepackage[utf8]{inputenc}
\usepackage[T1]{fontenc}      % pour que les accents soient correctement imprimés
\usepackage{graphicx}      % include this line if your document contains figures
\usepackage{natbib}        % required for bibliography
\usepackage{amsmath}               % great math stuff
\allowdisplaybreaks
\usepackage{amssymb}               % great math symbols
\usepackage{amsfonts}              % for blackboard bold, etc
\usepackage{amsbsy}% other symbols
\usepackage{bbm}
\usepackage{tikz}                  % to make drawing with TikZ
\usepackage{multicol}
\usepackage{graphicx}
\usepackage{threeparttable}
\usepackage{soul,derivative,stmaryrd,mathtools,enumerate,enumitem,mathtools}
\usepackage{comment}
\usepackage{algorithm}
\usepackage{algorithmicx}
\usepackage{algpseudocode}
\usepackage{enumitem}  % pour les sous-listes (a., b.)
\usepackage{ulem} % à mettre dans le préambule
\usepackage{hyperref}
\usetikzlibrary{arrows,positioning,shapes}
\usetikzlibrary{shapes.multipart}
\usetikzlibrary{arrows.meta}
\usetikzlibrary{decorations.pathreplacing}
\usetikzlibrary{arrows,intersections}
\usetikzlibrary{shapes,arrows}
\newcommand{\mbb}[1]{\mathbb{#1}}

\newcommand{\mrm}[1]{\mathrm{#1}}

\newcommand{\R}{\mbb{R}}

\newcommand{\I}{\mrm{I}}

\newcommand{\Pturb}{\mrm{P}_{\mrm{turb}}}
\newcommand{\n}{\mrm{n}}
\newcommand{\Xe}{\mrm{X}}

\newcommand{\Cb}{\mrm{C}_\mrm{b}}

\DeclareMathSymbol{\shortminus}{\mathbin}{AMSa}{"39}

\usepackage{subcaption} % dans le préambule
\author{%
 \Name{Perceval Beja-Battais} \Email{perceval.beja-battais@ens-paris-saclay.fr}\\
 \addr Université Paris Saclay, ENS Paris Saclay, CNRS, Centre Borelli, 91190, Gif-sur-Yvette, France
 \addr Framatome, Reactor Process Department, 92200, Courbevoie, France
 \AND
 \Name{Alain Grossetête} \Email{alain.grossetete@framatome.com}\\
 \addr Framatome, Reactor Process Department, 92200, Courbevoie,  France
 \AND
 \Name{Nicolas Vayatis} \Email{nicolas.vayatis@ens-paris-saclay.fr}\\
 \addr Université Paris Saclay, ENS Paris Saclay, CNRS, Centre Borelli, 91190, Gif-sur-Yvette, France
}

\begin{document}
%\raggedbottom

\maketitle

\begingroup
\renewcommand\thefootnote{}
\footnotetext{%
\textbf{Authors’ Version.}
This paper has been accepted and presented at ICAPP 2025.
Final version to appear in the proceedings published by SFEN.
\textit{This version includes minor stylistic and linguistic improvements; the technical content is unchanged.}%
}
\endgroup

%===============================================================================
\begin{abstract}
    In recent years, there has been an increasing need for Nuclear Power Plants (NPPs) to improve flexibility in order to match the rapid growth of renewable energies. The Operator Assistance Predictive System (OAPS) developed by Framatome addresses this problem through Model Predictive Control (MPC). In this work, we aim to improve MPC methods through data-driven simulation schemes. Thus, from a set of nonlinear stiff ordinary differential equations (ODEs), this paper introduces two surrogate models acting as alternative simulation schemes to enhance nuclear reactor core simulation. We show that both data-driven and physics-informed models can rapidly integrate complex dynamics, with a very low computational time (up to $1000\times$ time reduction).
\end{abstract}


\section{Introduction}

\subsection{Context}\label{sec:Context}

In the context of climate change, countries have claimed their will to replace fossil fuel power plants by greener, more renewable energy sources. However, these energy sources often are intermittent. In consequence, managing the electrical network becomes harder, as excess electricity cannot be stored on large scales. Hence, electricity producers try to match the electricity production to the real-time consumption~\citep{campagne2024leveraginggraphneuralnetworks}. For NPPs more specifically, flexibility represents a real challenge. Indeed, an overproduction of electricity would lead to both a waste of energy and an electricity market disruption~\citep{forsberg2013hybrid}. Nowadays, for recent NPPs, producers resort to load-following, i.e. adapting the production in real-time to match the needs from the electricity regulator. Load-following implies trade-offs, as chemical reactions inside core need to remain safely monitored. Moreover, economic trade-offs exist as the usage of control variables (e.g. boron) can be expansive. Such compromises naturally lead to the conception of controller minimizing costs while respecting these constraints. In the case of the OAPS System~\citep{dupre2025oaps}, the core is modeled through a set of nonlinear ODEs, and the trajectory optimization is addressed through Nonlinear MPC (NMPC) heuristics~\citep{dupre2021enhanced, dupre2022design, dupre2023conception}. This work is part of a continuous effort to improve this product.

\subsection{Related Work \& Motivations}

As described by~\cite{dupre2021enhanced}, the core can be represented as a stiff nonlinear ODE system. When integrating such systems, differential algebraic equations (DAE) is a natural representation in which fast-evolving variables are assumed to evolve instantly to their stationnary state~\citep{wanner1996solving}. Specialized solvers such as IDAS~\citep{serban2021user} or CVODES~\citep{cohen1996cvode} have been designed to integrate these types of dynamics. 

More recently, with the recent growth of computational resources, data-driven approaches for deterministic problems are of growing interest to address industrial systems~\citep{di2017learning, grigorescu2020survey, bertolini2021machine, usuga2020machine}. In many areas, deep learning models have outstripped the capabilities of previous existing models (e.g., Natural Language Processing~\citep{naveed2023comprehensive}, Computer Vision~\citep{voulodimos2018deep}...), sometimes at the expense of interpretability~\citep{shen2017deep}. For physics-based dynamical systems, recent ML models (e.g., Physics-Informed Machine Learning (PIML)~\citep{karniadakis2021physics}) provide cheaper and accurate simulations of a physical process, authorizing new ways to integrate Partial Differential Equations (PDEs). As an example, Physics-Informed Neural Networks (PINNs)~\citep{raissi2019physics} have been successfully implemented in many fields of application (e.g., fluid dynamics~\citep{cai2021physics, mao2020physics}, power systems~\citep{misyris2020physics}, geoscience~\citep{song2021wavefield}...). Nevertheless, though PINNs have shown promising results, it has been shown that they are subject to an important risk of overfitting~\citep{doumeche2023convergence}. 

For nuclear reactor core simulation, surrogate data-driven models have recently been implemented~\citep{bei2023surrogate, li2024new, antonello2023physics}. However, to the best of our knowledge, no machine learning surrogate dynamical system of the core has been developed for load-following purposes. In this paper, we show that the stiff component of the ODEs can be integrated through a PINN. We also demonstrate that we are able to integrate a set of stiff ODEs using XGBoost~\citep{ChenG16}. We believe that these approaches represent promising tools for MPC, as they allow us to carry a portable, fast and precise enough model. We believe that using such models could provide a close-to-optimal sequence of commands, or to warm-start a trusted simulation and optimization system, that is usually required in such sensitive industrial processes as nuclear energy. Note that the use of machine learning tools for MPC warm-start is an idea that has already been identified~\citep{klauvco2019machine}.
This paper is structured as follows: in Section~\ref{sec:background}, we recall the essential aspects of MPC and ML theories. In Section~\ref{sec:exps}, we present our two experiments on the industrial system. The first one, presented in Section~\ref{sec:Exp1}, consists in integrating the stiff component (neutron flux) of the set of ODEs through a PINN. The second one, presented in Section~\ref{sec:Exp2}, consists in integrating the stiff ODEs using XGBoost.  

\section{Background}\label{sec:background}

\renewcommand{\d}{\mathrm{d}}

\subsection{Notations}\label{sec:Notations}

Consider an ODE system represented by its state variable $x:\mathbb{R}^+ \to \mathbb{R}^d$ and dynamics $F:\mathbb{R}^d \to \mathbb{R}^d$ :
\begin{equation}\label{eq:dynamique_systeme}
\begin{aligned}
    \frac{\d x(t)}{\d t} &= F(x(t)).
\end{aligned}
\end{equation}

Assume we collected in the dataset $\mathcal{D}$ a large amount of simulations of this dynamical system. These simulations may reflect biases inherent to the simulation scheme. We will not, however, focus on this aspect in this paper.

\subsection{Numerical Simulation of ODEs}\label{subsec:stiff_ode}

While numerical simulation of complex ODEs has been heavily studied for decades~\citep{alexander1990solving, wanner1996solving}, certain multi-scale dynamics can still represent a challenge to integrate. Often, the stiffness of such systems will enforce the numerical scheme to take very small time steps in order to keep the simulation accurate. Dealing with those issues is automatically taken into account by specialized solvers~\citep{cohen1996cvode, serban2021user}. Nevertheless, those algorithms can be too long to evaluate for them to be integrated in an optimization pipeline. 

\subsection{Statistical Learning theory}\label{sec:Theory}

Assume access to a dataset of inputs and labels : $\mathcal{D} := \{(x_i, y_i), i\in\{1, \dots, N\}\}$. To replicate new data that the model would not have seen during its training, it is needed to randomly divide our dataset into a train set $\mathcal{D}_{\mathrm{train}}$ (typically containing $80\%$ of $\mathcal{D}$) and a test set $\mathcal{D}_{\mathrm{test}}$ (typically containing the last $20\%$). The test set is used to test the model's performances in close to real-life conditions. In the scope of this paper, we consider a regression setup. Given a function class $\mathcal{F}$ and a convex loss function $\ell$, we aim to find the function that best approximates the outputs from the inputs considering our training set $\mathcal{D}_\mathrm{train}$ : 
\begin{equation}\label{eq:stat_learning_regression_problem}
    \underset{f\in\mathcal{F}}{\min} \quad \mathbb{E}_{(x, y) \sim \mathcal{D}_{\mathrm{train}}}(\ell(y, f(x))).
\end{equation}

To solve this minimization problem, the expectancy operator is replaced by an empirical expectancy and a one-to-one mapping between $\mathcal{F}$ and a set of parameters in $\R^n$ is defined by $\phi:\alpha := (\alpha_1, \dots, \alpha_n) \mapsto f_\alpha \in \mathcal{F}$. The resulting optimization can be expressed as (assuming $\mathcal{D}_\mathrm{train}$ contains $N_\mathrm{train}$ samples) :
\begin{equation}\label{eq:emp_pb_with_param}
    \underset{\alpha \in \mathbb{R}^n}{\min} \quad \frac{1}{N_\mathrm{train}} \sum_{i=1}^{N_\mathrm{train}} \ell(y_i, f_\alpha(x_i)).
\end{equation}

Typically, minimizing over the train set such a data-dependent functional, accounting for data-fitting only, may lead to overfitting~\citep{shalev2014understanding}, i.e., very good estimations over the train set, but poor performances over the test set containing unseen data. To avoid this phenomenon, practical and theoretical estimation strategies resort to regularization~\citep{tian2022comprehensive}. Many ML methods are based on explicit regularization using penalties, while deep learning relies on implicit regularization~\citep{shalev2014understanding}.

\subsection{Artificial Neural Networks}\label{sec:NN}

For artificial neural networks (NN), $f_\alpha$ represents the neural network, while $\alpha$ represents its parameters at each layer. Formally speaking, a neural network is a successive composition of linear ($L$) and non-linear functions ($\sigma$)
\begin{equation}
    \mathcal{NN}(x) = \sigma_k \circ L_k \circ \dots \circ \sigma_0 \circ L_0(x),
\end{equation}
in which the parameters $\alpha = (\alpha_{ij})_{\substack{1\leq i \leq k \\ 1\leq j \leq n_k}}$ correspond to the coefficients of the linear combinations at each layer :
\begin{equation}
    L_i(x) = \sum_{j=1}^{n_k} \alpha_{ij} x_j.
\end{equation}

As the number of parameters inside the network can be very large, variants of stochastic gradient methods are generally used to perform the optimization from Eq.~\ref{eq:emp_pb_with_param} (e.g. Adam optimizer~\citep{kingma2014adam}). Different types of NN architectures exists (Multi-Layer Perceptrons, Recurrent Neural Networks, Convolutional Neural Networks...), each one being more or less adapted to specific problems. In this paper, we implement a Transformer-based neural network~\citep{vaswani2017attention}.

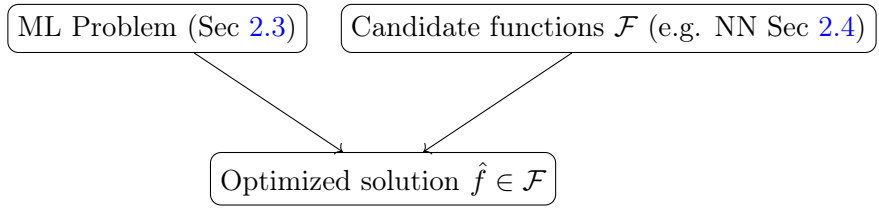
\begin{figure}[h!]
    \centering
    \begin{tikzpicture}
    % Nodes
    \node[draw, rounded corners] (ml) at (-3,2) { ML Problem (Sec~\ref{sec:Theory})};
    \node[draw, rounded corners] (f) at (3,2) { Candidate functions $\mathcal{F}$ (e.g. NN Sec~\ref{sec:NN})};
    \node[draw, rounded corners] (sol) at (0,0) { Optimized solution $\hat{f} \in \mathcal{F}$};

    % Arrows
    \draw[->] (ml) -- (sol);
    \draw[->] (f) -- (sol);
    \end{tikzpicture}
    \caption{Machine Learning Framework.}
    \label{fig:ML_framework}
\end{figure}

In the rest of this paper, we purposely drop the classical ML notation, i.e. $(x,y)$ being the input-output couple. To adapt to the physics-informed time series context, let $x(t+\d t) = f(x(t))$. The predictive ML model is assumed recursive i.e., the predictions over a whole horizon $T := N\d t$ are realized using previous outputs as inputs.

\subsection{Physics-Informed Neural Networks (PINNs)}\label{sec:PINNs}

\begin{figure}[h!]
    \centering
    \begin{tikzpicture}
    % Cercles des ensembles
    \fill[blue!30, opacity=0.6] (-2,0) ellipse (3 and 2);
    \fill[red!30, opacity=0.6] (2,0) ellipse (3 and 2);
    \fill[green!50, opacity=0.6] (-1.05,-0.65) circle (1.2);
    \draw[thick] (-2,0) ellipse (3 and 2) node[left=0.4] {ML Problems};
    \draw[thick] (2,0) ellipse (3 and 2) node[right=0.5] {PI Problems};

    % Sous-ensemble NN
    \draw[thick, dashed] (-1.05,-0.65) circle (1.2) node[left=0.2] {NN};

    % Intersection ML Models \ Physics-Informed Problems = PIML
    \node at (0,0.8) {PIML};

    % Intersection NN \ Physics-Informed Problems = PINN
    \node at (-0.4,-0.5) {PINN};
    \end{tikzpicture}
    \caption{Relationship Between Machine Learning (ML) and Physics-Informed (PI) Problems. The green circle illustrates the subset of ML Problems that can be solved through the usage of neural networks.}
    \label{fig:venn_nn}
\end{figure}
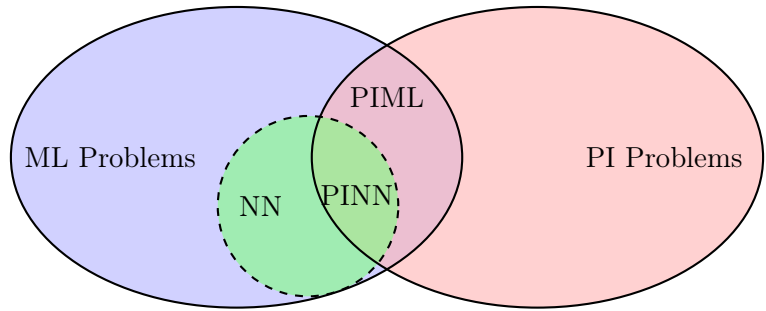

In recent years, interest has grown around the use of deep neural networks for physical simulation, especially to solve ODEs and PDEs~\citep{raissi2019physics, cai2021physics, sahli2020physics, nguyen2022physics, cuomo2022scientific}. PINNs are neural networks that are trained by extending the formulation after the classic data-based optimization problem from Eq.~\eqref{eq:emp_pb_with_param} (see Fig.~\ref{fig:venn_nn}). 

Let $\Omega \subset \R^d$ and $x(t, \chi)$ be a solution of an arbitrary PDE :
\begin{equation}\label{eq:PDE}
\forall t > 0, \forall \chi \in \Omega, \left\{
\begin{aligned}
    F(t, \chi, x, \partial_t x,\partial_\chi x, \dots) &= 0, \\
    x(0, \chi) &= x_0(\chi).
\end{aligned}
\right.
\end{equation}

Let $\mathcal{D} := \{(t^\mathrm{data}_i, \chi^\mathrm{data}_i, x_i), i \in\{1, \dots, N_\mathrm{data}\}\}$ a dataset containing simulations of the physical phenomenon. Assume knowledge of a part of the equations that make up the system $F(t, \chi, x, \partial_t x, \partial_\chi x, \dots) = 0$ and let $f(t,\chi)$ the prediction at point $(t, \chi)$ of the NN. Defining a set of collocation points $(t_i^\mathrm{colloc}, \chi_i^\mathrm{colloc}) \in \mathbb{R}^+_* \times \Omega$, the NN is implicitly being guided to solve the known PDEs at the collocation points.  In practice, this is done by combining the data loss (Eq.~\eqref{eq:data_loss}) with a physics-informed loss (Eq.~\eqref{eq:physics_loss} and Eq.~\eqref{eq:boundary_loss}):

\begin{equation}\label{eq:data_loss}
    L_D(f) := \frac{1}{N_\mathrm{data}} \sum_{i=1}^{N_\mathrm{data}} \|f(t^\mathrm{data}_i, \chi^\mathrm{data}_i) - x_i\|^2, \quad \text{(Data Loss)}
\end{equation}

\begin{equation}\label{eq:physics_loss}
    L_\phi(f) := \frac{1}{N_\mathrm{colloc}} \sum_{i=1}^{N_\mathrm{colloc}} \|F(t^\mathrm{colloc}_i, \chi^\mathrm{colloc}_i, f, \partial_t f, \partial_\chi f, \dots) \|^2, \quad \text{(Dynamics Loss)}
\end{equation}
\begin{equation}\label{eq:boundary_loss}
    L_{\partial \chi}(f) := \frac{1}{N_\mathrm{bound}}\sum_{i=1}^{N_\mathrm{bound}} \|f(0, \chi^\mathrm{bound}_i) - x_0(\chi^\mathrm{bound}_i)\|^2. \quad \text{(Boundary Loss)}
\end{equation}

Given a set of functions $\mathcal{F}$, the new optimization problem is given by
\begin{equation} \label{eq:PINN_opti_pb}
    \underset{f \in \mathcal{F}}{\min} \quad \alpha_D L_D(f) + \alpha_\phi L_\phi(f) + \alpha_{\partial \chi}L_{\partial \chi}(f),
\end{equation}
where $\alpha_D, \alpha_\phi, \alpha_{\partial \chi}$ correspond a weighting between the physics equations and the relative closeness to the data points. One can numerically solve this optimization problem by using similar stochastic gradient techniques to the previous part (e.g. SGD or Adam~\citep{kingma2014adam}).

In the rest of this paper, we neglect the boundary loss from Eq.~\eqref{eq:boundary_loss} as we consider we know the initial point $x(0)$ from current measures. As a consequence, we fix $\alpha_{\partial \chi} = 0$ and focus on the dynamics loss (Eq.~\eqref{eq:physics_loss}).

\subsection{Transformers}\label{subsec:Transformers}

\begin{figure}[h!]
    \centering
    \includegraphics[width=0.3\linewidth]{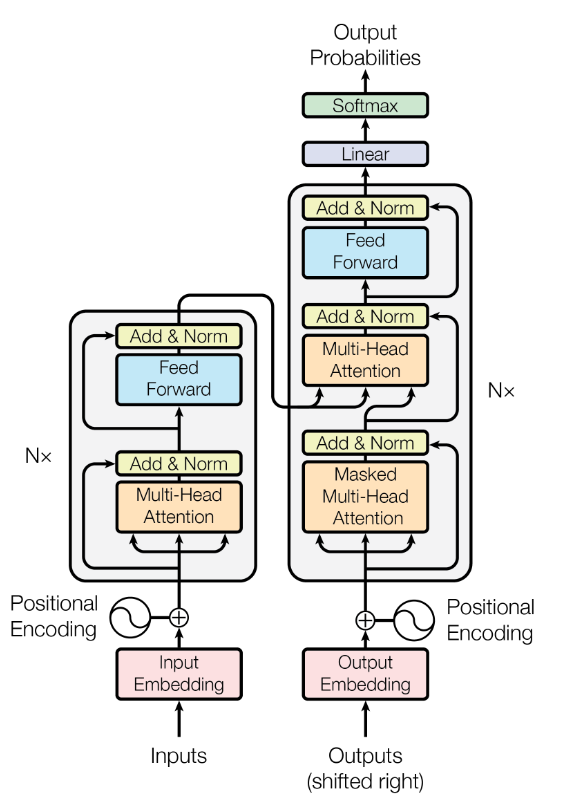}
    \caption{Vanilla architecture of a transformer, from~\cite{vaswani2017attention}.}
    \label{fig:transformer}
\end{figure}

Transformers (Fig.~\ref{fig:transformer}) were introduced by~\cite{vaswani2017attention} in order to better capture complex dependencies across the elements of a sequence. This new architecture has shown great success in many fields of application of deep neural networks~\citep{lin2022survey, islam2023comprehensive} (e.g., computer vision~\citep{amjoud2023object}, time-series analysis~\citep{ahmed2023transformers}, natural language processing~\citep{bracsoveanu2020visualizing}, reinforcement learning~\citep{li2023survey}...). 

Transformers capture spatial and temporal dependencies through a measure called \textit{attention}. Formally speaking, attention is a mapping from a triplet $(Q, K, V)$ to an output $A(Q, K, V)$ where $Q$ is a query and $D = (K, V)$ a key-value dictionary. A compatibility score between the query $Q$ and the keys $K_1, \dots, K_n$ of the dictionary is computed, often through a dot-product, and these scores are then used to weight the reference values of the dictionary. For example, the \textit{Scaled Dot-Product Attention} described by~\cite{vaswani2017attention} is computed via the operation :
\begin{equation}\label{eq:scaled_dot_prod_attention}
    \mathrm{Attention}(Q, K, V) = \mathrm{Softmax}\left(\frac{QK^T}{\sqrt{d_K}}\right)V
\end{equation}
where $K, V$ respectively stand for the keys and values of $D$, and $d_K$ stands for the dimension in which lays the keys from $K$.

This computation can be effectively parallelized in order to compute multiple attention heads, depending on the queries and dictionaries given in input. For instance, one may project a query and dictionary onto a lower-dimensional subspaces via several mappings $(Q, K, V) \mapsto (QW_i^Q, KW_i^K, VW_i^V)$ and then combine the obtained attention heads to compute a \textit{Multi-Head Attention}~\citep{vaswani2017attention}, the idea being to capture different perspectives and relationships of the data. 

\subsection{Sequence-to-Sequence Learning}

Sequence-to-Sequence learning~\citep{sutskever2014sequence} (Seq2Seq) encompasses the ensemble of machine learning models that aim to predict a sequence from another sequence (typically for translation in Natural Language Processing) by encoding a known sequence of tokens $x_1, \dots, x_T$ (e.g. words, or values for numerical time series) onto a context vector $h_T$. This vector is then given as input to a decoder trained to predict the output sequence $y_1, \dots, y_{T'}$. Though Seq2Seq was initially designed for Recurrent Neural Networks (RNN) encoder and decoder architectures, it has shown great success being implemented on Transformers architecture, thanks to the attention mechanism~\citep{li2023transformer, lu2021context}.

\subsection{eXtreme Gradient BOOSTing (XGBoost)}

XGBoost is a machine learning algorithm developed by~\cite{ChenG16}. It has proven to be a popular algorithm for time series forecasting~\citep{zhang2021time, paliari2021comparison}. Similarly to other boosting methods, XGBoost creates a sequence of weak estimators~\citep{schapire1999brief} (e.g. decision trees) that successively correct the errors from the previous weak estimators through a reweighting of the data from the train set. The boosting method then outputs a prediction by combining the prediction from the weak estimators.

\section{Experiments}\label{sec:exps}
In this section, we describe our two experiments. In the first one we design a fast numerical integration scheme for the stiff component of our system. This is done with a Physics-Informed Transformer. In the second one, we show that XGBoost achieves high accuracy as an integrator of the stiff ODEs. 

\subsection{ODE System}\label{subsec:ODEsystem}

\newcommand{\Cbrcp}{C_{\mathrm{b, rcp}}}
\newcommand{\Cbrcv}{C_{\mathrm{b, rcv}}}
\newcommand{\Cbinj}{C_{\mathrm{b, inj}}}
\newcommand{\Tbf}{T_{\mathrm{cl}}}
\newcommand{\T}{\mathrm{T}}
\newcommand{\Xbank}{\mathrm{X}_{\mathrm{bank}}}
\newcommand{\GF}{\mathrm{GF4}}

We model the reactor core as described in~\cite{dupre2021enhanced}:
\begin{itemize}
    \item the 1D core has $n_z$ vertical meshes, each mesh characterized by its iodine concentration $\I_i$, xenon concentration $\Xe_i$ and neutron density $\mathrm{n}_i$,
    \item the control rods $\Xbank$ interacts with the cold leg temperature $\Tbf$ through the French N4 power plants temperature regulation~\citep{dupre2021enhanced, dupre2022design},
    \item the boron concentration $\Cb$ in the core is assumed to be constant throughout the experiments.
\end{itemize}

The ODE system is given by (considering $\Pturb(t)$ as an input control variable, and $\Cb(t) = \Cb$) :

\begin{equation} \label{eq:DAE_system}
    \begin{aligned}
        \frac{\d \n}{\d t} \!&=\! F_\n(\n(t), \Xe(t), \T(t), \Xbank(t))  \\
        \frac{\d \I(t)}{\d t} \!&=\! F_\I(\I(t), \n(t)) \\
        \frac{\d \Xe(t)}{\d t} \!&=\! F_\Xe(\Xe(t), \I(t), \n(t)) \\
        \frac{\d \Tbf(t)}{\d t} \!&=\! F_{\Tbf}(\n(t), \Pturb(t)) \\
        \frac{\d \Xbank(t)}{\d t} \!&=\! F_{\Xbank}(\Pturb(t), T(t))
    \end{aligned}
\end{equation}

In all the following, let $x(t) := (\n(t), \I(t), \Xe(t), \Tbf(t), \Xbank(t)) \in \mathbb{R}^{N}$ (with $N = 3n_z + 2$ in our case) and $n_z = 6$.

\subsection{Experiment 1 : Predicting the neutron flux through a Physics-Informed Transformer}\label{sec:Exp1}

Integrating the neutron flux inside the reactor core can be a difficult task because of the neutron dynamics which is very fast compared to any other dynamic. Thus, the challenge is to integrate a highly nonlinear and stiff component of the ODE defined in Sec.~\ref{subsec:stiff_ode}. 
To do so, we trained a Seq2Seq Physics-Informed Transformer the following way :
\begin{itemize}
    \item the inputs are a past sequence of state trajectory $x(t_0), \dots, x(t-\d t)$ and the non-stiff components of $x(t)$,
    \item the target output is the stiff component $\n(t)$,
    \item the model interacts with a classic integrator (e.g. Euler scheme) that outputs the non-stiff components of $x(t+\d t)$ in order to recursively generate a trajectory for the system,
    \item the physics-informed loss is computed through residuals at each timestamp of the predicted sequence, and data loss is computed with the gap between the reference trajectory and the predicted trajectory.
\end{itemize}

The choice of a Transformer architecture was motivated through a quick benchmark. For a comparable number of learnable parameters, MLPs were not able to properly tame the nonlinear dynamics, whereas Recurrent Neural Network architectures such as LSTM have shown poorer performance and more long-term error propagation. We trained the PINN over 315 transients of 24h and test it over 79 transients of 24h with $\d t = 60$ s, and over a combination of the data and physical loss in order to generate physically feasible solutions. We give in Table~\ref{tab:Exp1_results} quantitative results of our experiment.

\begin{table}[h!]
    \centering
    \begin{tabular}{|c|c|c|}
        \hline
         Average MSE/min (\%NP) & PI Residual Error (Eq.~\eqref{eq:physics_loss}) & Computational time (for 24h)\\
        \hline
         $0.13 \pm 0.11$ & $(2 \pm 1) \cdot 10^{-6}$ & $(3.8 \pm 0.1) \cdot 10^{-3}$ s \\
        \hline
    \end{tabular}
    \caption{Results of Experiment 1 on the test set. While the alternative integration method is a lot faster, the mean error is reasonably low. As a comparison, a reference solver (IDAS~\citep{serban2021user}) takes on average 5 seconds over the same horizon.}
    \label{tab:Exp1_results}
\end{table}

\begin{figure}[h!]
    \centering
    \begin{minipage}{0.5\textwidth} % Ajustez la largeur si nécessaire
        \includegraphics[width=\linewidth, viewport=0 0 600 600, clip]{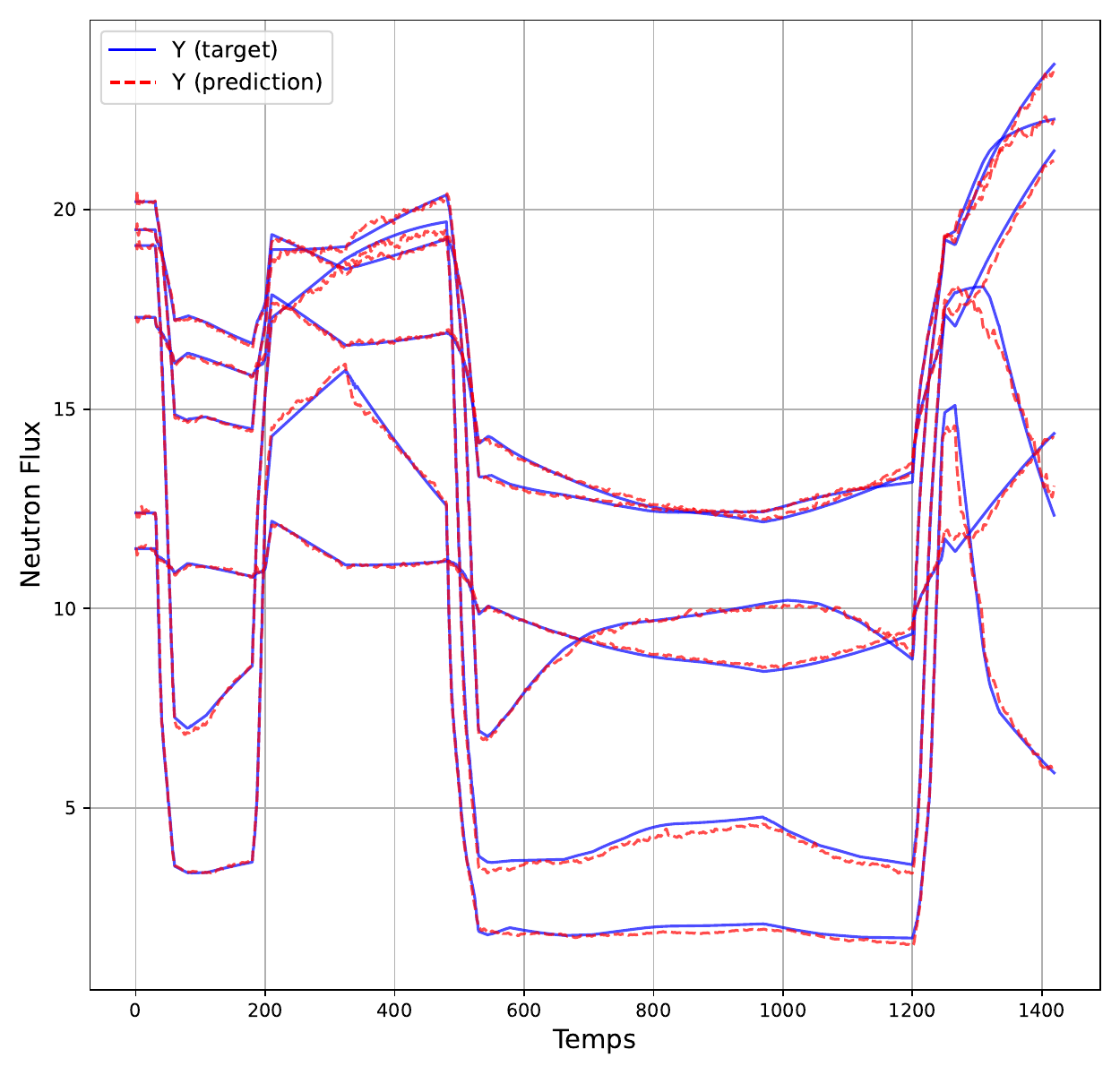}
    \end{minipage}
    \caption{Comparison between the neutron flux integrated by the described method and a reference (IDAS) for the given power transient.  Time ($x$-axis) is displayed in minutes.}
    \label{fig:Exp1_Exemple}
\end{figure}

In Figure~\ref{fig:Exp1_Exemple}, we display a comparison between the predictions of $\n(t), \dots, \n(t+N\d t)$ by the described method and a reference solver (IDAS~\citep{serban2021user}). In this example, the turbine power decreased, at $t = 30$ min, from 100\% NP to 70\% NP at a 1\% NP/min rate. It goes back up to 100\% NP at a 1\% NP/min rate at $t = 180$ min.  At $t = 480$ min, the power goes back down to 50\% NP and up to 100\% NP at $t = 1200$ min at the same rate. Throughout the 24 hours, the boron concentration $\Cb$ is assumed to be fixed at 1296 ppm.

The model turns out to achieve reasonable accuracy considering the little computation time to obtain a 24h trajectory. Even though the dynamics are not perfectly tamed, a usual solver such as IDAS or CVODES~\citep{cohen1996cvode} would have taken a considerable amount of small time steps ($\d t \leq 10^{-3}$ s) to integrate the ODE, rendering a 5 s computation time on the same hardware.

\subsection{Experiment 2 : Predicting the stiff system using XGBoost}\label{sec:Exp2}

In this experiment, we design a model able to do recursive long term predictions (i.e. 24h). We set the model with 100 base estimators, with in input the future $N$ steps of turbine power $\Pturb(t), \dots, \Pturb(t + N\d t)$ as well as the current state $x(t)$. It predicts both the stiff and non-stiff components of $x(t+\d t), \dots, x(t+N\d t)$.

\begin{figure}[H]
%\begin{figure}[]
    \vspace{-4em}
    \centering
    % Première sous-figure
    \begin{minipage}[b]{0.49\textwidth}
        \includegraphics[width=\textwidth, viewport=0 0 700 750]{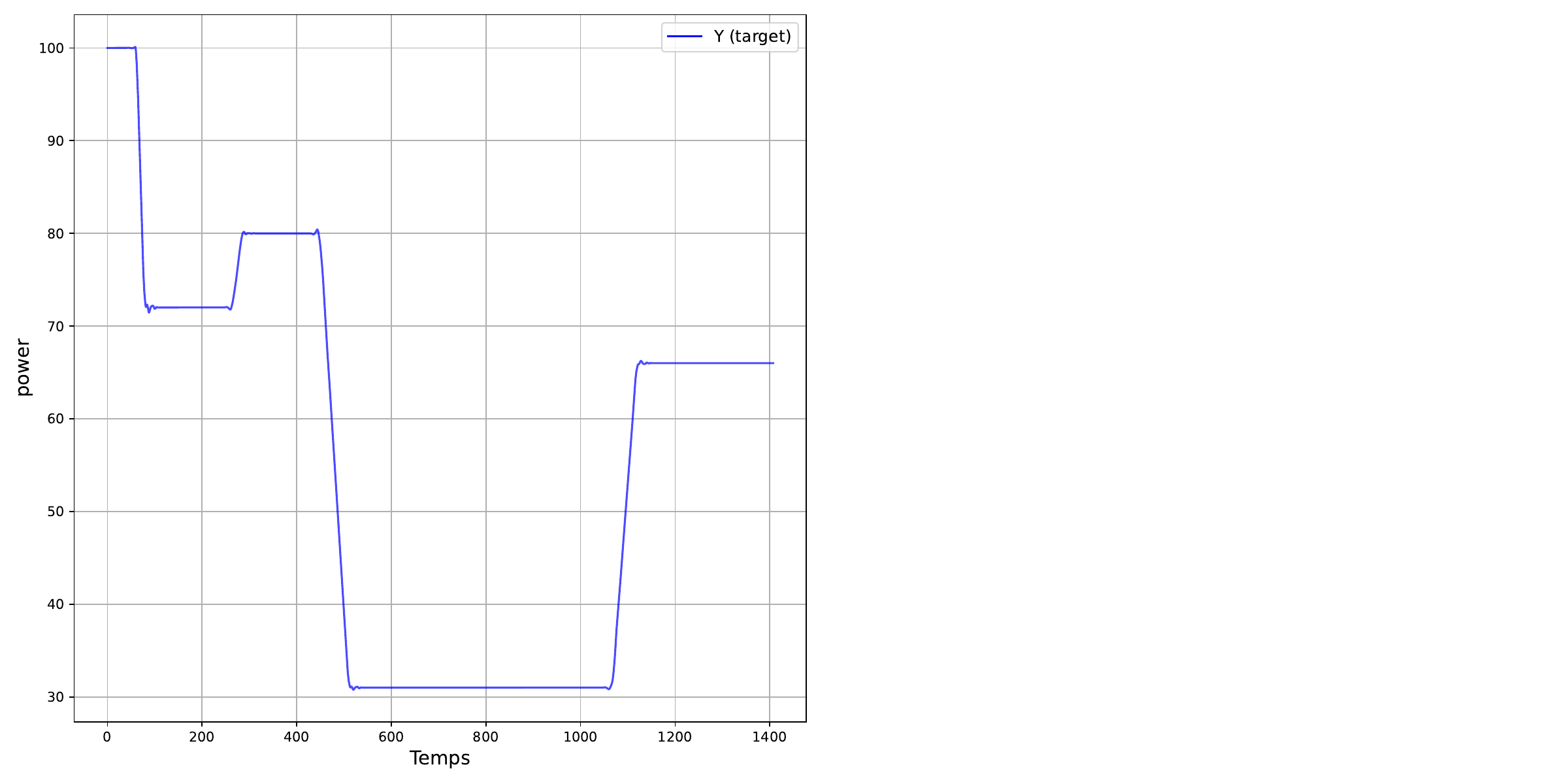}
    \end{minipage}
    \hfill
    \begin{minipage}[b]{0.49\textwidth}
        \includegraphics[width=\textwidth, viewport=0 0 700 750]{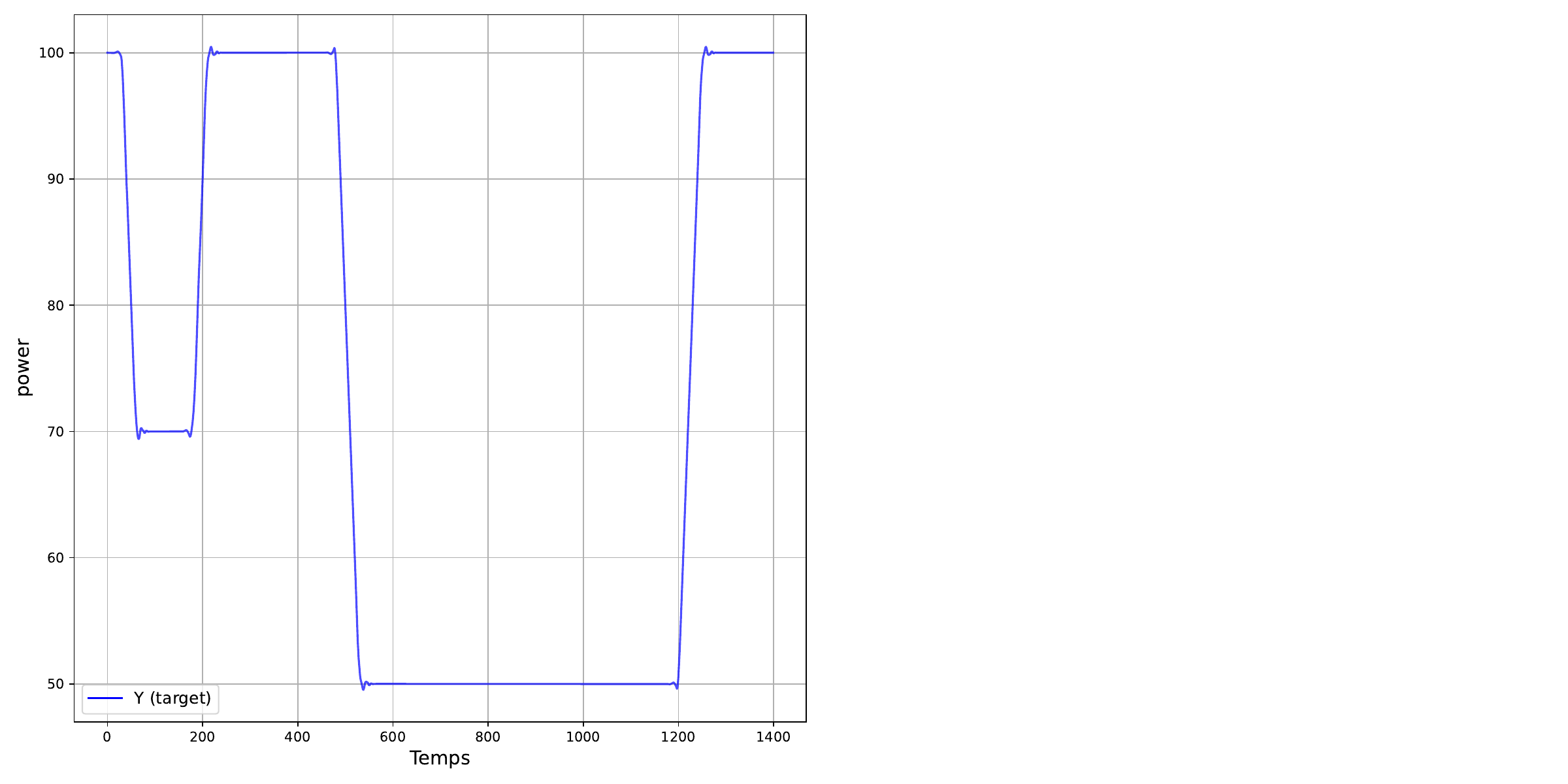}
    \end{minipage}
    \caption{Input power $\Pturb(t)$ for the two examples for the XGBoost model.}
    %\caption{Experiment 2 : Input power $\Pturb(t)$ for the two examples for the XGBoost model.}
    \label{fig:Exp2_Exemple1}
%\end{figure}
\end{figure}

\begin{figure}
    \centering
    \begin{minipage}[b]{\textwidth}
        \includegraphics[width=\textwidth]{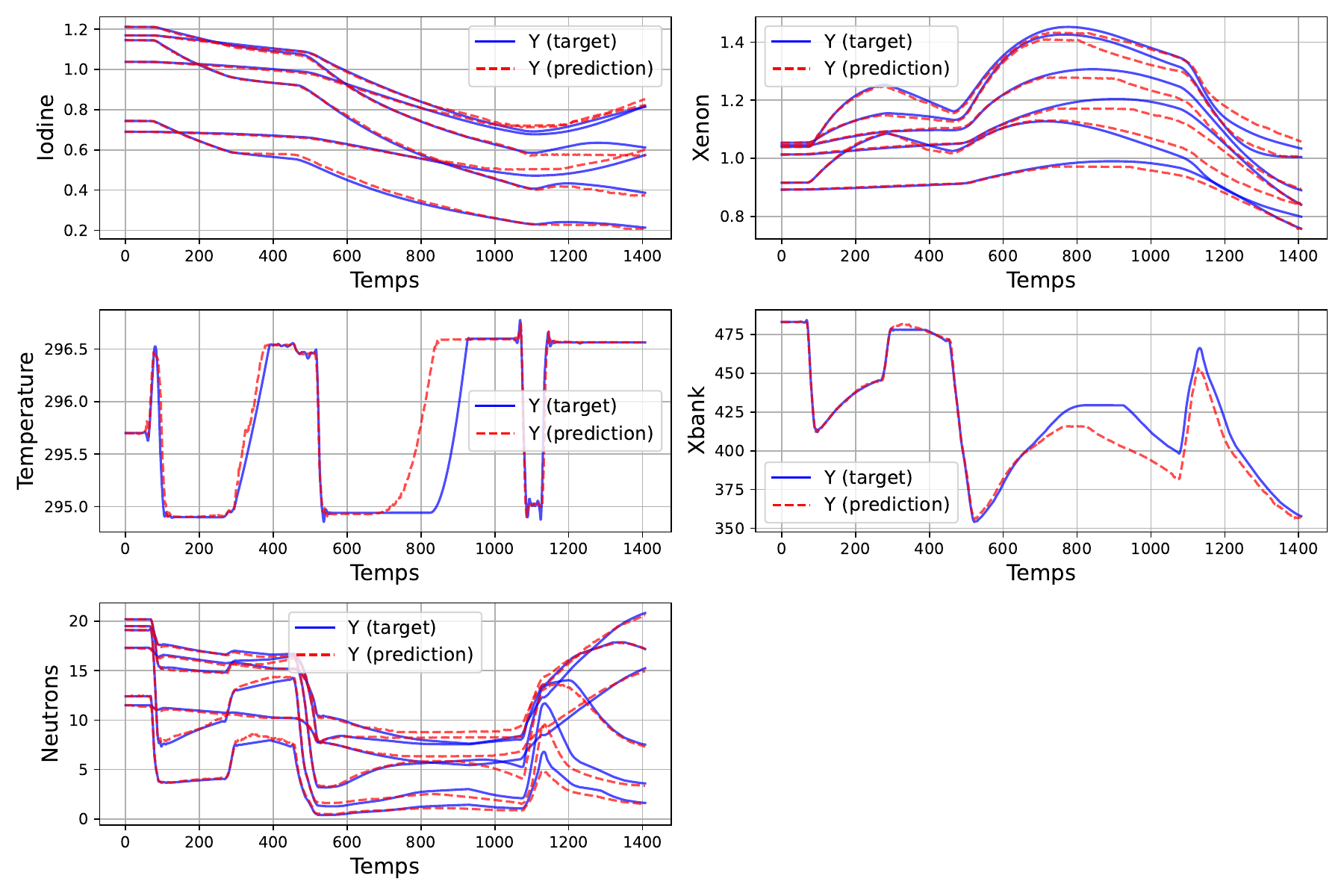}
    \end{minipage}
    \vfill
    % Deuxième sous-figure
    \begin{minipage}[b]{\textwidth}
        \includegraphics[width=\textwidth]{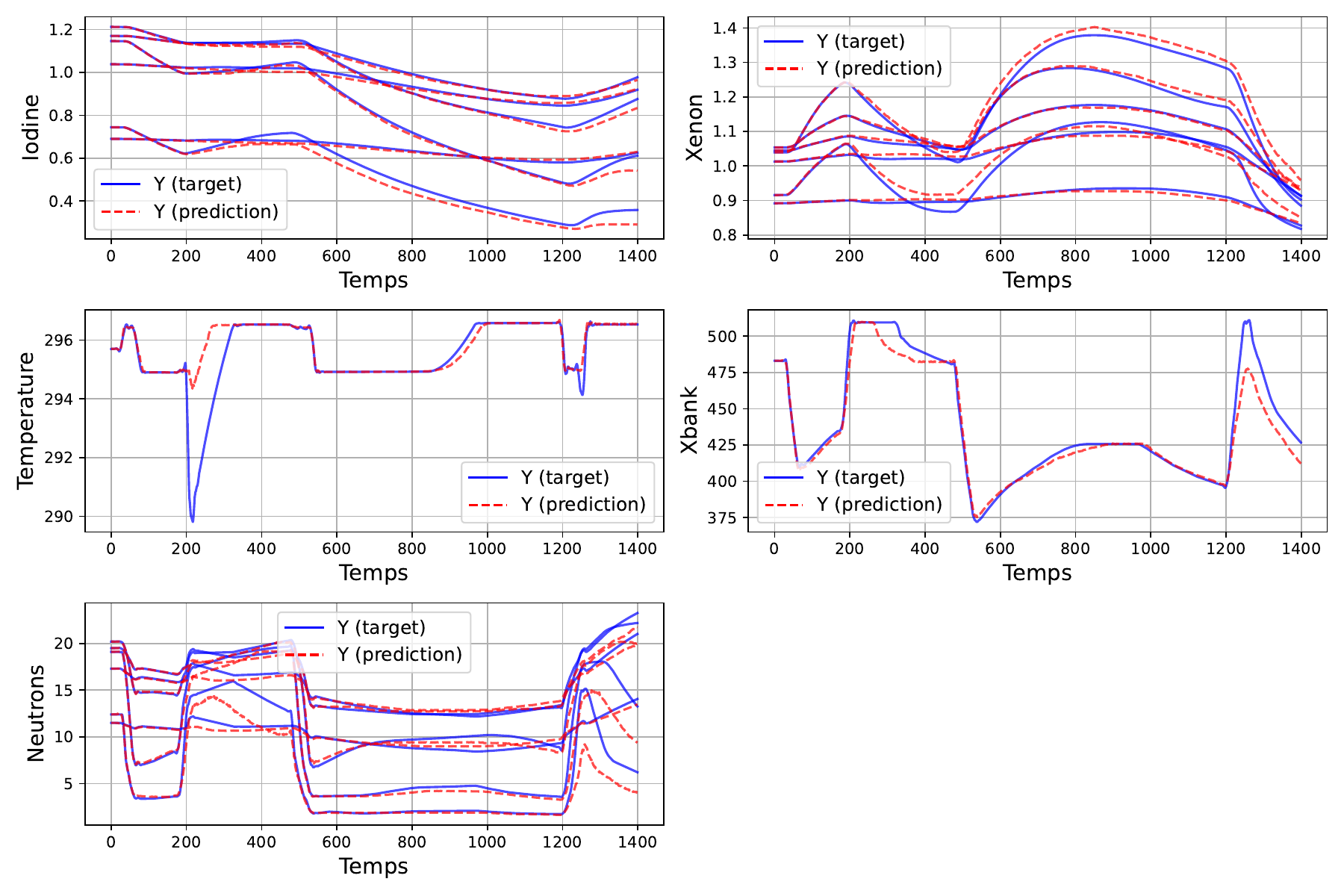}
    \end{minipage}

    \caption{Comparison between the predictions and the ground truth on the two selected examples. Temperature stands for $\Tbf(t)$.}
    \label{fig:Exp2_Example2}
\end{figure}

To predict a longer trajectory, the model can be called recursively, i.e. $x(t+N\d t)$ can be used an input of the next prediction. In our experimental setup, we set $N=10$. In practice, the lower $N$, the higher training samples can be used for the same amount of data (by cutting the trajectories into $x(t) \mapsto x(t+\d t), \dots, x(t+N \d t)$ for $t = 0, \dots, T - N \d t$). The results are shown in Tab.~\ref{tab:Exp2_results}, Fig.~\ref{fig:Exp2_Exemple1} and Fig.~\ref{fig:Exp2_Example2}. 
\begin{table}
    \centering
    \begin{tabular}{|c|c|c|}
        \hline
        Variable & Scaled MSE on 24h\\
        \hline
         $\I(t)$ & $2.7\pm 4.5$\\
         $\Xe(t)$ & $6.2\pm 15$\\
         $\Tbf(t)$ & $0.4\pm 0.4$\\
         $\Xbank(t)$ & $1.1\pm 1.3$\\
         $\n(t)$ & $4.0\pm 6.9$\\
         \hline
         $x(t)$ (Overall) & $14\pm 26$\\
        \hline
    \end{tabular}
    \caption{Results on the test set. Scaled MSE stands for MSE over the normalized space in which all components are between 0 and 1 (to be able to compare variables that do not have the same scale (e.g. $\I, \Xe$ and $\Tbf, \Xbank$)). As expected, the stiff component $\n(t)$ is among the most difficult variables to predict. Moreover, long term error propagation appears over the coupled system $\I(t), \Xe(t)$ leading to higher error on those "easy" variables.}
    \label{tab:Exp2_results}
\end{table}

Beyond the quantitative results, what stands out of this experiment is the physical coherence of the model: when a shift appears on a variable, it quite coherently impacts the other variables. For instance, in Fig.~\ref{fig:Exp2_Exemple1}, around $t = 800$ min, a shift happens over $\Tbf$ and $\Xbank$. The predicted correlation between the variables is physically correct, even though it led to a significant shift with respect to the reference. This traduces the model abilities to learn the main physical links between variables. 

However, compared to the previous experiment, the predictions are not as fast to obtain, and computation time are similar to a traditional solver. A deeper reflection should be led on the choice of the model to tame the system dynamics while keeping a low computational cost.

% Première figure

% Deuxième figure

\section{Conclusion}

Data-driven approaches hold great promise for integrating complex dynamics that would usually require a very high amount of computations due to the stiffness of a system. Through the two presented experiments, we show that data-driven approaches can be leveraged for nuclear reactor core simulation. We believe that these models, merged with other traditional MPC methods, represent promising tools for optimal control.

Using data-driven forecasting models for simulation to an optimal control problem such as the one from~\cite{dupre2022design} is a challenge that will need to be addressed. We believe models such as PINNs can turn out to be a powerful tool if used to find a suboptimal trajectory that will then be used as a warmstart for the classic optimization pipeline.

\bibliography{l4dc2026-sample}

\end{document}